\documentclass[lettersize,jMS-GSnal]{IEEEtran}
\usepackage{amsmath,amsfonts}
\usepackage{algorithmic}
\usepackage{algorithm}
\usepackage{array}
\usepackage[caption=false,font=normalsize,labelfont=sf,textfont=sf]{subfig}
\usepackage{textcomp}
\usepackage{stfloats}
\usepackage{url}
\usepackage{verbatim}
\usepackage{graphicx}
\usepackage{cite}
\usepackage{colortbl} 
\usepackage{xcolor} 
\usepackage{amssymb}
\usepackage{multirow}
\usepackage{booktabs}
\usepackage{arydshln}
\usepackage{amsmath}
\usepackage{color}
\usepackage{threeparttable}
\usepackage{etoolbox}
\usepackage{xcolor}
\definecolor{lightred}{RGB}{255,199,206}
\definecolor{lightpurple}{RGB}{229,227,250}
\newcommand{\M}{Scale-GS}
\makeatletter
\patchcmd{\@makecaption}
  {\scshape}
  {}
  {}
  {}
\makeatother

\begin{document}

\title{Scale-GS: Efficient Scalable Gaussian Splatting via Redundancy-filtering Training on Streaming Content }
\author{Jiayu~Yang, Weijian~Su, Songqian~Zhang, Yuqi~Han, 
Jinli~Suo,
Qiang~Zhang,~\IEEEmembership{Senior Member,~IEEE}

\thanks{Jiayu~Yang, Weijian~Su, Songqian~Zhang, Yuqi~Han, and Qiang~Zhang are with the School of Computer Science and Technology, Dalian University of Technology, Dalian 116024, China, and are also with Key Laboratory of Social Computing and Cognitive Intelligence (Dalian University of Technology), Ministry of Education, Dalian, 116024 China. Jinli~Suo is with the Department of Automation, Tsinghua University, Beijing 100084, China.}
\thanks{
(\emph{Corresponding authors: Yuqi~Han; yqhanscst@dlut.edu.cn.})}
}

\maketitle

\begin{abstract}
3D Gaussian Splatting (3DGS) enables high-fidelity real-time rendering, a key requirement for immersive applications. However, the extension of 3DGS to dynamic scenes remains limitations on the substantial data volume of dense Gaussians and the prolonged training time required for each frame.
This paper presents \M, a scalable Gaussian Splatting framework designed for efficient training in streaming tasks. Specifically, Gaussian spheres are hierarchically organized by scale within an anchor-based structure.
Coarser-level Gaussians represent the low-resolution structure of the scene, while finer-level Gaussians, responsible for detailed high-fidelity rendering, are selectively activated by the coarser-level Gaussians.
To further reduce computational overhead, we introduce a hybrid deformation and spawning strategy that models motion of inter-frame through Gaussian deformation and triggers Gaussian spawning to characterize wide-range motion. Additionally, a bidirectional adaptive masking mechanism enhances training efficiency by removing static regions and prioritizing informative viewpoints. Extensive experiments demonstrate that \M~ achieves superior visual quality while significantly reducing training time compared to state-of-the-art methods.

\end{abstract}
\vspace{-3mm}

\section{Introduction}
The rapid advancement of 3D Gaussian Splatting (3DGS)~\cite{kerbl20233d} has significantly reshaped the domain of real-time 3D rendering. In particular, the introduction of Gaussian training methods designed for dynamic scenes~\cite{huang2024sc,tong2025rigid,fan2025rgavatar,fan2025fov,yang2024deformable,li2024spacetime,wu20244d,yan20244d} has greatly enhanced the feasibility of 3D streaming applications, including virtual reality (VR), augmented reality (AR), and immersive telepresence systems. By explicitly representing scenes with differentiable Gaussians, these methods facilitate real-time rendering—capabilities that are critical for interactive applications demanding low-latency visual feedback.
However, as computational demands scale sharply with scene and temporal complexity, the training time for dynamic scenes—ranging from tens of minutes to hours—conflicts with the low-latency requirements of real-time streaming.

The primary reason for the slow training time in Gaussian Splatting stems from redundant computations involving Gaussian spheres.  The standard 3D Gaussian splatting methods process each frame independently, thereby incurring repetitive calculations on predominantly static Gaussians due to the limited extent of dynamic regions. Although recently some research partitions the scene into static and dynamic components to focus computational resources on dynamic Gaussians, the overall volume remains substantial. Consequently, significant overlapping computations occur across spatially and temporally adjacent regions, leading to inefficient resource usage and bottlenecks that hinder real-time performance.

\begin{figure*}[!t]
  \centering
  \includegraphics[width=0.9\textwidth]{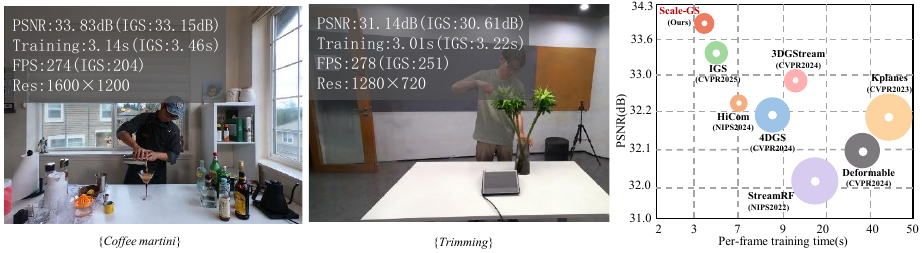}
  \caption{The proposed~\M~ under dynamic scene achieves best rendering quality with the shortest training time. The left figures show results of our~\M~on N3DV Coffee martini and MeetRoom Trimming datasets, where ``Res'' indicates video resolution. The right figure is tested on the N3DV dataset, where the radius of the circle corresponds to the average storage per frame and the method in the top left corner demonstrates the best performance.}
  \label{teaser}
  \vspace{-3mm}
\end{figure*}

We observe that the size and number of Gaussian spheres to represent the 3D scene vary significantly depending on scene complexity. For example, textureless planar regions can be effectively modeled by a small number of large Gaussian spheres, whereas highly textured areas necessitate dozens or even hundreds of smaller Gaussians. The larger Gaussian spheres often contribute more significantly to scene representation and therefore warrant higher training priority. Inspired from scalable video coding, this work proposes~\M, a scalable Gaussian splatting framework aimed at mitigating redundancy in 3D spatial representation, thereby enabling accelerated training. Specifically, Gaussian spheres are organized by scale, with each level independently trained under the viewpoints at corresponding resolutions. For each frame, large scale Gaussians are first optimized using low-resolution views to approximate the scene. Upon convergence, a triggering criterion evaluates whether small scale Gaussians should be activated for refinement, thereby reducing the redundant Gaussian spheres optimization. As shown in Fig.~\ref{teaser}, the proposed method achieves high rendering quality with short training time compared to the SOTA algorithm. The framework conducts a coarse-to-fine principle where the training at each level of scale is triggered by the preceding level of scale, thereby filtering redundant training of irrelevant Gaussians. 

To enable efficient training for streaming content, we introduce a hybrid strategy that combines deformation and spawning to infer dynamic changes in the current frame based on the preceding frame. Generally, the Gaussian deformation~\cite{yan20244d,yang2024deformable} models the motion of Gaussian sphere, but insufficient for capturing newly appearing objects or wide-range motion. In contrast, Gaussian spawning~\cite{sun20243dgstream,shao2023tensor4d} introduces new Gaussians to fine-tune dynamic regions but requires considerably longer training time. To balance the trade-off, the hybrid strategy first applies deformation to model inter-frame motion and then determines, based on the training outcome, whether spawning should be triggered for finer-grained refinement. Specifically, under the~\M~framework, if the deformation of a determined scale fails to adequately represent the dynamic,  either the emergence of new content or the motion of smaller-scale Gaussians happens. Thus, the deformation result triggers new Gaussians spawning at the current scale and deformation at the next scale. This sequential activation ensures Gaussians are progressively introduced at locations with actual dynamics along with increasingly finer scales, enabling efficient and high-fidelity temporal GS representation.

To enhance training efficiency, we propose a bidirectional adaptive masking mechanism that simultaneously suppresses static regions and selects informative training viewpoints. The forward masking component detects dynamic and static anchors via inter-frame change analysis, where pixel-wise differences between consecutive frames are back-projected to estimate motion patterns. For backward camera viewpoint selection, we define a relevance score between projected dynamic anchors and camera fields of view, further weighted by directional factors that prioritize orthogonal or novel viewpoints. The top-ranked views, as determined by the relevance score, are selected to form the active viewpoint set. This bidirectional masking mechanism reduces computational redundancy caused by uninformative viewpoints and facilitates accurate reconstruction of dynamic scenes.

Comprehensive experiments conducted on three challenging real-world datasets—NV3D, MeetRoom, and Google Immersive—demonstrate the superior performance of the proposed framework across multiple evaluation metrics.  Qualitative comparisons show that our method reconstructs significantly sharper fine-grained details, particularly in complex scenarios involving human interactions, dynamic phenomena such as flames, and intricate textures. Furthermore, experimental results demonstrate that~\M~not only improves visual quality but also outperforms current state-of-the-art methods in both training and rendering time. These findings validate the effectiveness of~\M, which prioritizes more important Gaussian spheres and improves the average training efficiency.

The main contributions of this work are as follows:
\begin{enumerate}
    \item We propose~\M, a scalable GS framework performing redundancy-filtering training on streaming content to improve the efficiency. The ~\M~achieves the most efficient training compared to the existing Gaussian training methods on streaming content.
    \item The~\M~integrates a hybrid deformation-spawning Gaussian training strategy that prioritizes large-scale Gaussians and selectively activating finer ones to reduce redundancy in 3D scene representation while preserving high-fidelity dynamic representations.
    \item Extensive evaluations show that~\M~achieves superior efficiency–quality trade-offs for streaming novel view synthesis, improving visual quality, reducing training time, and supporting real-time rendering.
\end{enumerate}

In the following, we first introduce the research related to the~\M, including the novel view synthesis for static scene and videography at Sec.~\ref{sec2}. Later we thoroughly present the detail of the method at Sec.~\ref{sec3}. The qualitative and quantitative experimental results are exhibits at Sec.~\ref{sec4}. Finally, we draw the conclusion and propose the future work at Sec.~\ref{sec5}.

\section{Related work}
\label{sec2}
In this section, we separately review research on implicit and explicit novel view synthesis methods for both static and dynamic scenes. These studies primarily focus on improving visual quality and enhancing training efficiency.
\subsection{Novel View Synthesis for Static Scenes}
Early novel view synthesis methods predominantly rely on geometric interpolation, with approaches such as the Lumigraph \cite{gortler2023lumigraph} and Light Field rendering \cite{meng20203d,levoy2023light}, laying the groundwork through advanced interpolation techniques applied to densely sampled input images.

Neural Radiance Fields (NeRF) \cite{mildenhall2021nerf} introduces a breakthrough in photorealistic view synthesis by modeling scene radiance through implicit neural representations using multi-layer perceptrons. This innovation has spurred extensive research aiming at overcoming NeRF's inherent limitations across various dimensions. Key efforts include accelerating training procedures \cite{chen2022tensorf,hu2024culling,muller2022instant,fridovich2022plenoxels}, achieving real-time rendering performance \cite{chen2023mobilenerf,ye2024real,yu2021plenoctrees}, improving synthesis fidelity in complex scenes \cite{barron2021mip,barron2022mip,ye2024real}, and enhancing robustness under sparse input conditions \cite{lu2024relightable,wimbauer2023behind,yu2021pixelnerf}. However, the computational overhead inherent in NeRF’s volume rendering paradigm—which requires numerous neural network computation per frame—presents significant challenges in balancing training efficiency, rendering speed, and visual fidelity.

To address these limitations, Kerbl et al. \cite{kerbl20233d} proposes 3DGS, which leverages explicit 3D Gaussian primitives combined with differentiable rasterization-based rendering to enable real-time, high-quality view synthesis. The 3DGS inspires a broad range of research efforts exploring various aspects of Gaussian-based scene representations. Some studies focus on enhancing rendering fidelity \cite{lu2024scaffold,ren2024octree,chen2024pgsr}, while others aim to improve geometric precision and accuracy \cite{yu2024gaussian,lin2025hybridgs}. Furthermore, considerable efforts are devoted to developing compression techniques to mitigate storage overhead \cite{chen2024hac,lu2024scaffold,lee2024compact,niedermayr2024compressed,li2025mpgs}. Recent studies investigate the joint optimization of camera parameters alongside Gaussian field estimation \cite{fu2024colmap}, as well as the extension of Gaussian splatting to broader 3D content generation tasks \cite{tang2024lgm,tang2025ivr,zou2024triplane}. Despite the 3DGS in static scenes achieves high-quality rendering, the development of an on-demand training framework for 3DGS remains an open challenge.

Generalization is introduced as an effective strategy to enhance inference speed. Recent developments in NeRF methodologies \cite{chen2021mvsnerf,johari2022geonerf,yu2021pixelnerf} and 3DGS\cite{szymanowicz2024splatter,zhang2025transplat,zheng2024gps} focus on generalizable reconstruction networks trained on large-scale datasets. Specifically, PixelSplat \cite{charatan2024pixelsplat} uses Transformers to encode features and decode Gaussian primitive parameters. DepthSplat \cite{xu2025depthsplat} leverages monocular depth to recover 3D details from sparse inputs. Other frameworks \cite{chen2024mvsplat,liu2024mvsgaussian,zhang2024gs} combine Transformer or Multi-View Stereo (MVS) \cite{yao2018mvsnet} methods to build geometric cost volumes, enabling real-time generalization.
However, due to the insufficient diversity of available 3D datasets, the generalization performance of these methods remains to be further improved.
\vspace{-3mm}
\subsection{Novel View Synthesis for Dynamic Scenes}
Novel view synthesis for dynamic scenes naturally extends static models, with early approaches building on NeRF \cite{cao2023hexplane,fridovich2023k,guo2023forward,lin2023high,liu2022devrf,pumarola2021d,shao2023tensor4d} and 3DGS \cite{kerbl20233d}, leveraging their efficient rendering capabilities for dynamic scene reconstruction.
While Gaussian-based methods \cite{huang2024sc,tong2025rigid,fan2025rgavatar,fan2025fov,yang2024deformable,li2024spacetime,wu20244d,yan20244d} learn temporal attributes to model dynamic scenes as unified representations and improve reconstruction quality, their requirement to load all data simultaneously results in high memory usage, limiting their feasibility for long-sequence streaming.

To address these challenges, streaming-based methods, such as ReRF \cite{wang2023neural}, NeRFPlayer \cite{song2023nerfplayer}, and StreamRF \cite{li2022streaming} reformulate dynamic scene reconstruction as an online problem. Moreover, 3DGStream \cite{sun20243dgstream} utilizes Gaussian-based representations combined with Neural Transformation Caches to model inter-frame motion, though it still requires over 10 seconds per frame. HiCoM \cite{gao2024hicom} introduces an online reconstruction pipeline for multi-view video streams employing perturbation-based smoothing for robust initialization and hierarchical motion coherence mechanisms. Instant Gaussian Stream (IGS) \cite{yan2025instant} proposes enables single-pass motion computation guided by keyframes, reducing error accumulation and achieving reconstruction times around 4 seconds per frame. Alternative frame-tracking methods \cite{guo2024motion,luiten2024dynamic} track Gaussian evolution across frames, supporting streaming protocols but incurring substantial per-frame data overhead.

In contrast to existing methods that require full sequence processing or incur significant per-frame optimization overhead, we propose~\M, a scalable Gaussian splatting framework for efficient streaming rendering. By constructing multi-scale Gaussian representations combined with selective training,~\M~enables on-the-fly novel view synthesis while preserving high rendering quality.

\section{Method}
\label{sec3}
\begin{figure*}[!t]
	\centering
	\includegraphics*[width=.95\linewidth]{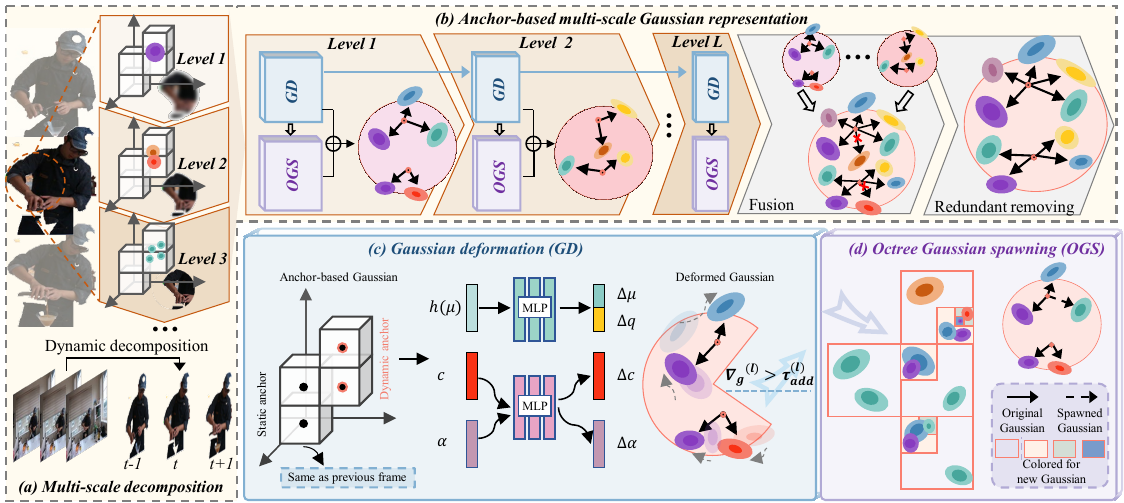}
	\caption{The framework of ~\M. (a) The multi-scale decomposition across different resolution levels, where finer scales capture increasingly detailed scene dynamics. (b) Anchor-based multi-scale Gaussian representation. After completing training at each level, all scales are combined, followed by redundant Gaussians removing. (c-d) The hybrid deformation and spawning Gaussian optimization. (c) Gaussian deformation module that models temporal changes through anchor-guided MLPs, (b) Octree Gaussian Spawning that adaptively adds new Gaussians based on Octree subdivision.} 
    \label{framework}
    \vspace{-3mm}
\end{figure*}

In this section, we first introduce the preliminaries in Sec.~\ref{3.1}. Later we present the key pipeline of~\M. The framework of~\M~ is presented as in Fig.~\ref{framework}. After decompositing the dynamic part, \M~follows an anchor-based multi-scale Gaussian representation~(Sec.~\ref{3.2}) as its core framework. We apply hybrid deformation-spawning Gaussian optimization~(Sec.~\ref{3.3}) to model inter-frame motion. When deformation is insufficient to capture dynamics, we activate the next scale and selectively spawn new Gaussians.  Once all scales have converged, redundant Gaussians are pruned to optimize the representation~(Sec.~\ref{3.4}). In addition, ~\M~employs bidirectional bidirectional adaptive masking~(Sec.~\ref{3.5}) to identify dynamic anchors and select informative viewpoints.
\vspace{-4mm}


\subsection{Preliminaries}
\label{3.1}
3DGS uses a dense set of Gaussian spheres to represent the whole space, and renders viewpoints via differentiable splatting combined with tile-based rasterization of these Gaussian components. For each Gaussian sphere $i$, the expectation position $\mu_i$ and variance $\Sigma_i$ determine the formation of the Gaussian $G_i(x)$. 
The point $x$ on the Gaussian sphere $G_i(x)$ is noted as
\begin{equation}
G_i(x) = \exp\left( -\frac{1}{2} (x - \mu_i)^{\top} \Sigma_i^{-1} (x - \mu_i) \right), \label{eq:1}
\end{equation}
where $\Sigma_i$ is composed of a rotation matrix $R_i \in \mathbb{R}^{3 \times 3}$ and a diagonal scale matrix $S_i \in \mathbb{R}^{3 \times 3}$, denoted as $\Sigma_i = R_i S_i S_i^T R_i^T$. The color and opacity of Gaussian sphere $i$ are denoted as $c_i$ and $\alpha_i$.

We use $x'$ to define the 2D projection pixel position, and the pixel value \( C \in \mathbb{R}^3 \) is rendered via \(\alpha\)-composite blending. Specifically, we assume the light ray projected onto a pixel $x'$ intersects with $N$ Gaussian surfaces along its path. The color $C(x')$ is defined as
\begin{equation}
C(x') = \sum_{i \in N} c_i \alpha_i \prod_{j=1}^{i-1} (1 - \alpha_j), \label{eq:2}
\end{equation}
where $N$ Gaussians are sorted from near to far and $\alpha_i$ signifies the opacity of Gaussian $G_i(x)$.
With the differentiable rasterizer, all attributes of the 3D Gaussians become learnable and can be directly optimized in an end-to-end manner through training view reconstruction.

To enable structural rendering in the Gaussian splatting model, we introduce an anchor-based mechanism\cite{lu2024scaffold,chen2024hac}. The 3D space is uniformly partitioned into multiple voxels, with each voxel assigned a dedicated anchor responsible for managing all Gaussian primitives within its region. 

The anchor-based framework is initialized by voxelizing the sparse point cloud generated from Structure-from-Motion (SfM) pipelines. Let $v$ denote the index of a specific anchor, and $V$ represent the complete set of anchors across the entire space. Each $v \in V$ corresponds to a local context feature \( \hat{f}_v \), a 3D scaling factor \( c_v \), and \( k \) learnable offsets \( O_v \in \mathbb{R}^{k \times 3} \) to identify the attribute of the anchor $v$ and $k$ corresponding Gaussians (indexed from $0$ to $k-1$).
Specifically, given the location $x_v$ of anchor $v$, the positions of Gaussians winthin the anchor $v$ are derived as
\begin{equation}
    \{\mu_0, \ldots, \mu_{k-1}\} = x_v + \{O_0, \ldots, O_{k-1}\} \cdot c_v. \label{eq:3}
\end{equation}

Since the Gaussians associated with the same anchor share similar attributes, the other factors of Gaussians could be predicted by lightweight MLPs \( F_\alpha, F_c, F_q, F_s \) taking the anchor attributes as input. Specifically, we denote the relative viewing distance from anchor $v$ to the camera as $\delta_{v}$ and the viewing direction from anchor $v$ to the camera as $\vec{d}_{v}$. Taking opacity prediction as an example, the opacity values of all $k$ Gaussians within anchor $v$ are derived as 
\begin{equation}
    \{\alpha_0, \ldots, \alpha_{k-1}\} = F_\alpha(\hat{f}_v, \delta_{v}, \vec{d}_{v}). \label{eq:4}
\end{equation}

The color, rotation, and scale predictions follow similar formulations using their respective MLPs $F_c$, $F_q$, and $F_s$.

By partitioning the whole space into voxels and assigning an anchor to each voxel, the anchor-based approach facilitates localized organization and efficient indexing of Gaussian distributions, significantly reducing the overhead of traversing and computing irrelevant Gaussians. Moreover, the neural Gaussians, which infer all Gaussian factors from the anchor attribute improves the efficiency of training.

\subsection{Anchor-based Multi-Scale Gaussian Representation}
\label{3.2}


Variations in texture detail of the 3D scene lead to Gaussian representations of differing granularity. To improve computational efficiency, we introduce a multi-scale Gaussian optimization framework, drawing inspiration from scalable video encoding\cite{mizuho2024reduction,groth2023wavelet}. In this framework, coarse-scale Gaussians are first optimized using low-resolution inputs, followed by the refinement of fine-scale Gaussians guided by high-resolution images. 
We separate static and dynamic regions and observe the dynamic parts. When inter-frame variations arise, optimization starts at the coarse scale and progressively refines finer scales, ensuring global structures and local details, as shown in Fig.~\ref{framework}(a).

We define a multi-scale structure with \( L \) levels of scale and \( l \) corresponds to a specific level.
The $M$ training viewpoint at time $t$ with original resolution as $I_{0,t},...,I_{M-1,t}$. The viewpoint resolution at the level $l$ is denoted as $I^{l}_{0,t},...,I^l_{M-1,t}$.
At each level \( l \), the corresponding set of Gaussians \( \mathcal{G}^{(l)} \) is supervised by viewpoints $I^{l}_{0,t},...,I^l_{M-1,t}$.
Considering that variations in Gaussian representations between temporally adjacent frames of the same scene are limited, the distribution learned from the initial frame can reasonably approximate that of all frames. Thus, we initialize the 3DGS on the first frame to estimate the scale of each Gaussian.
We define the  maximum scale $s_\text{max}^{(0)}$, minimum $s_\text{min}^{(0)}$, and mean $s_\text{mean}^{(0)}$ at each level $l$. 

Given that Gaussian spheres with larger scales encode less fine-grained detail, we employ a binary partitioning strategy to divide the scale space. Specifically, once the scale range for a given level is established,  the subsequent level is recursively defined within the finer half of the current level’s range.
Specifically, we assume the size of level $l$ is defined as \([s_\text{min}^{(l)}, s_\text{max}^{(l)}]\). If the level $l+1$ is required, the scale of level $l$ is revised to \([s_\text{mean}^{(l)}, s_\text{max}^{(l)}]\) and the the scale of level $l+1$ is revised to \([s_\text{min}^{(l)}, s_\text{mean}^{(l)}]\), i.e,
$[s_\text{min}^{(l+1)}, s_\text{max}^{(l+1)}] \leftarrow [s_\text{min}^{(l)}, s_\text{mean}^{(l)}], \quad [s_\text{min}^{(l)}, s_\text{max}^{(l)}] \leftarrow [s_\text{mean}^{(l)}, s_\text{max}^{(l)}]$.
This ensures that higher-indexed levels (higher resolution) consistently receive smaller Gaussian scale ranges. 

We introduce a clamp function to ensure that Gaussians at each level are constrained within their designated scale ranges, which is represented as \( \text{clamp}(,) \), to enforce both upper and lower bounds on the scale of the Gaussians, i.e.,
\begin{equation}
s_i = \text{clamp}(s_\text{min}^{(l)}, s_\text{max}^{(l)},s_i), \text{if } s_i\in l. \label{eq:5}
\end{equation}

According to Eq.~(5), scale parameters at each level are constrained within their designated ranges, ensuring that coarser and finer Gaussians are optimized independently.

\vspace{-3mm}

\subsection{Hybrid Deformation-spawning Gaussian Optimization}
\label{3.3}

\begin{figure}[t]
	\centering
	\includegraphics*[width=1\linewidth]{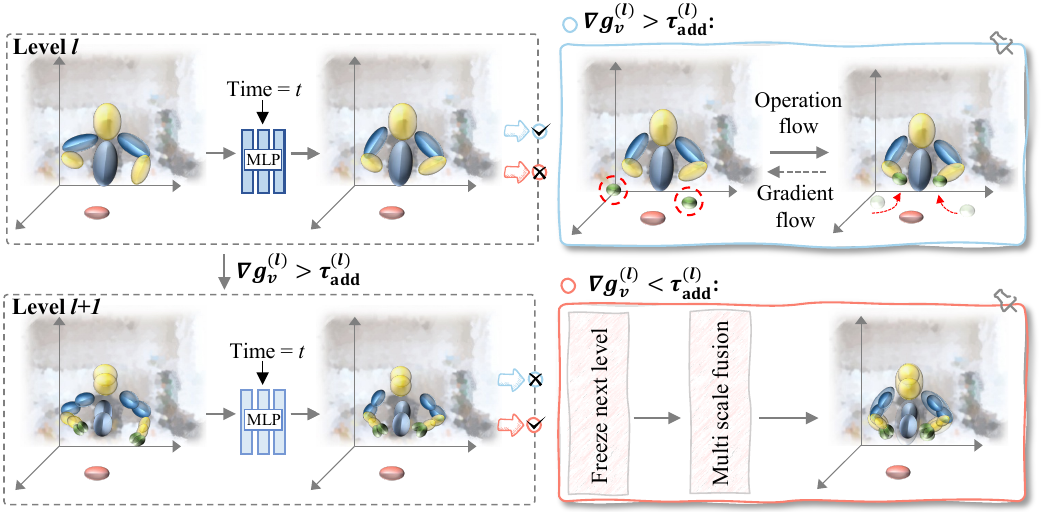}
	\caption{The detail of hybrid deformation-spawning strategy across multi-scale levels. At level $l$, Gaussians undergo temporal deformation via MLPs. When the average gradient exceeds the threshold , the next level $l+1$ is activated for finer-grained optimization. Meanwhile, within each subspace at level $l$, when the mean gradient exceeds the threshold, new Gaussians are spawned within that subspace. Conversely, when the average gradient is less than the threshold, the hierarchical progression stops and multi-scale fusion is performed to integrate representations across all active levels.}
        \label{detail}
        \vspace{-4mm}
\end{figure}

As the scene changes over time, existing methods rely on either deformation, which lacks expressiveness for complex dynamics, or spawning, which is computationally inefficient due to the need for many new Gaussians. To overcome the limitations, we propose a hybrid approach that combines deformation and spawning policy. After performing deformation at scale level $l$, unresolved dynamics—either from higher-resolution variations or newly emerged components—are addressed by guided spawning at the current scale and activating deformation at the $l+1$, as shown in Fig.~\ref{framework}(b).

We use the anchor $v$ as an example to present the following description. For scale level $l$, the deformation process is processed through two lightweight Multi-Layer Perceptrons (MLPs) to model the dynamics of each Gaussian from $t-1$ to $t$. We adopt a decoupled design to separately model geometry and appearance. Geometric deformation leverages spatial context via hash encoding to capture 3D structure, while photometric changes—reflecting intrinsic material properties—are processed directly without spatial encoding. The framework of Gaussian deformation is presented as Fig.~\ref{framework}(c). At each level $l$, an geometric deformation MLP ($\mathrm{MLP_g}$) takes the previous frame’s center position $\mu^{(l)}_{i,t-1}$ as input, which is encoded via multi-scale hash encoding $h(\mu^{(l)}_{i,t-1})$ to capture both local and global spatial context.
The $\mathrm{MLP_g}$ processes the hash-encoded position to predict geometric changes
\begin{equation}
\Delta\mu^{(l)}_i, \Delta q^{(l)}_i = \mathrm{MLP_g}(h(\mu^{(l)}_{i,t-1})). \label{eq:6}
\end{equation}
The output is a 7-dimensional vector, where the first 3 dimensions represent the position offset $\Delta\mu^{(l)}_i$ and the last 4 dimensions denote the quaternion increment $\Delta q^{(l)}_i$.

The appearance deformation MLP ($\mathrm{MLP_a}$) receives the original color $c^{(l)}_{i,t-1}$ and opacity $\alpha^{(l)}_{i,t-1}$ as inputs, and predicts the updated values $c^{(l)}_{i,t}$ and $\alpha^{(l)}_{i,t}$, i.e.,
\begin{equation}
c^{(l)}_{i,t}, \alpha^{(l)}_{i,t} = \mathrm{MLP_a}(c^{(l)}_{i,t-1}, \alpha^{(l)}_{i,t-1}). \label{eq:7}
\end{equation}

Overall, we model temporal change of all Gaussian attributes at level $l$ via residual updates from the previous frame
\begin{equation}
\begin{aligned}
\theta_i^{(l,t)} &= \theta_i^{(l,t-1)} + \Delta \theta_i^{(l)},
\quad \theta \in \{\boldsymbol{\mu}, \Sigma, \alpha, \mathbf{c}\},\\
q^{(l)}_{i,t} &= \mathrm{norm}(q^{(l)}_{i,t-1}) \cdot \mathrm{norm}(\Delta q^{(l)}_i), \label{eq:8}
\end{aligned}
\end{equation}
where $\mathrm{norm}(\cdot)$ indicates quaternion normalization to ensure unit quaternion constraints for valid rotations. 

The rendering of scale level $l$ follows the volume rendering approach where each level processes its own Gaussians independently
\begin{equation}
C^{(l)}(x') = \sum_{i \in N}c_{i,t}^{(l)}\alpha_{i,t}^{(l)}\prod_{j=1}^{i-1}(1-\alpha_{j,t}^{(l)}). \label{eq:9}
\end{equation}

The reconstruction loss at scale level $l$ is computed with respect to the corresponding resolution images, including an $\ell_1$ loss and a structural similarity loss to enforce perceptual fidelity.
\begin{equation}
\mathcal{L}^{(l)} =\mathcal{L}_1^{(l)}(I^l_{n,t},\hat C^l_{n,t})+ \lambda_{\text{SSIM}} \mathcal{L}_{\text{SSIM}}^{(l)}(I^l_{n,t},\hat C^l_{n,t}), \label{eq:10}
\end{equation}
where $\hat C_{n,t}$ indicates the volume rendering result of viewpoint $n$ at time $t$ with Gaussian scale level $l$.

For each anchor $v$ at level $l$, we compute the average gradients of constituent Gaussians over $d$ training iterations, denoted as $\nabla g_v^{(l)}$, denoted as
\begin{equation}
\nabla g_v^{(l)} = \frac{1}{|G_v^l|} \sum_{i \in G_v^l} \left\|\frac{\partial \mathcal{L}}{\partial \theta_i^{(l)}}\right\|, \label{eq:11}
\end{equation}
where $G_v^l$ represents the set of Gaussians belonging to anchor $v$ at scale level $l$.

We define a level-specific gradient thresholds~\cite{lu2024scaffold}, denoted as 
\begin{equation}
\tau_{\text{add}}^{(l)} = \frac{Vol}{4^{l-1}}, \label{eq:12}
\end{equation}
where $Vol$ indicates the volume size of each anchor and the threshold decreases exponentially with scale level $l$, guiding progressively finer control at higher resolution levels. If  $\nabla g_v^{(l)}>\tau_{\text{add}}^{(l)}$ after deformation at $l$, suggesting that the resolution of $l$ is not sufficient to capture the underlying dynamic changes, the~\M~framework triggers octree Gaussian Spawning at $l$ and deformation at $l+1$. The detail of hybrid deformation-spawning strategy across multi-scale is represented as Fig.~\ref{detail}.

\noindent \textbf{Octree Gaussian Spawning of Scale $l$.} As shown in Fig.~\ref{framework}(d), for the Spawning of level $l$, Gaussians are spawned at predefined anchor locations, followed by an optimization phase to adapt the newly added Gaussians to the dynamic scene. After training, these Gaussians are associated with their corresponding anchors and carried forward for overfitting in the subsequent frame.

We construct an octree-based Gaussian representation to model dynamics with minimal Gaussian usage. The octree structure enables adaptive allocation of Gaussian guided by gradient information. 
We refer to each region partitioned by the octree within an anchor space as a subspace. At each level $l$, we compute the mean gradient $\nabla g_{\text{v}}^{(l)}$ of all Gaussians within each subspace. If $\nabla g_{\text{v}}^{(l)}>\tau_{\text{add}}^{(l)}$, a fixed number of Gaussians are randomly assigned within the subspace, which is then recursively subdivided. This process continues until all subspaces fall below the threshold or the spatial resolution reaches one-thousandth of the original domain.

\noindent \textbf{Gaussian Deformation of Scale $l+1$.}
To enable finer-grained deformation modeling, the optimization process activates Gaussians at the next resolution level $l{+}1$. These Gaussians are inherited from frame $t{-}1$ and share the same anchor structure as level $l$, but operate at a higher spatial resolution through hierarchical refinement.
\vspace{-3mm}
\subsection{Redundant Gaussian Removing}
\label{3.4}

As the hybrid deformation-spawning strategy introduces additional Gaussians in each frame, a redundant Gaussian removing step is applied after all scales converging to prevent uncontrolled growth in the number of Gaussian over time. 
For each Gaussian $i$ managed by anchor $v$, we define a 1-D optimizable mask $M_{v,i}$.
The \( M_{v,i} \) is passed through a sigmoid function \( \sigma(\cdot) \) to ensure differentiability.
When \( M_{v,i} \) is large, \( \sigma(M_{v,i}) \rightarrow 1 \); conversely, when \( M_{v,i} \) is small, \( \sigma(M_{v,i}) \rightarrow 0 \), effectively removing the corresponding Gaussian.

For rendering pixels covered by anchor $v$, the projection color after masking is defined as:
\begin{equation}
    C(x') = \sum_{i \in N}\sigma(M_{v,i})c_{i,t}\alpha_{i,t}\prod_{j=1}^{i-1}(1-\sigma(M_{v,i})\alpha_{j,t}). \label{eq:13}
\end{equation}

When the sigmoid value \( \sigma(M_{v,i}) \) approaches zero, the corresponding Gaussian \( i \) under anchor \( v \) is excluded from the volume rendering process, regardless of the scale level from which it originated. Thus, the loss function is then employed to optimize the Gaussians, filtering out redundant ones. The total loss combines the reconstruction error from the multi-scale fused rendering with a sparsity regularization term, i.e.,
\begin{equation}
\mathcal{L} = \sum_{l=1}^L \mathcal{L}^{(l)} + \lambda_{\text{r}} \sum_{v \in V} \sum_{i=1}^N \sigma(M_{v,i}), \label{eq:14}
\end{equation}
where $\mathcal{L}^{(l)}$ represents the rendering loss on each scale, and $\lambda_{\text{r}}$ controls the redundancy removing regularization. The redundancy removing term reduces redundant Gaussians by penalizing non-zero mask values, thereby promoting a compact representation across all anchors and scale levels.

\subsection{Bidirectional Adaptive Masking}
\label{3.5}
We propose a bidirectional adaptive masking mechanism that selects dynamic spatial anchors and informative camera views for each training frame, serving as a preprocessing step to reduce redundant computation in static regions and mitigate supervision from views with limited marginal utility.

\begin{algorithm}[!t]
\caption{The optimization of~\M~}
\label{alg:dynamic_update}
\begin{algorithmic}[1]
\REQUIRE Prior Gaussians $\mathcal{G}_{t-1}$, frame $I_t$, scale levels $L$, thresholds $\tau_{\text{add}}^{(l)}$.
\ENSURE Updated $\mathcal{G}_t$;
\STATE Initialize: $\mathcal{G}_t \leftarrow \mathcal{G}_{t-1}$;
\STATE Apply bidirectional adaptive masking (\ref{3.5}) to identify dynamic anchors $V^{\text{dyn}}$ using Eq.(\ref{eq:16});
\FOR{each level $l = 1$ to $L$}
    \STATE Apply scale constraints using Eq.(\ref{eq:5});
    \FOR{each dynamic anchor $v \in V^{\text{dyn}}$}
        \STATE Apply deformation using Eq.(\ref{eq:6}-\ref{eq:7}): geometric and appearance MLPs;
        \STATE Compute reconstruction loss using Eq.(\ref{eq:10});
        \STATE Compute gradient $\nabla g_v^{(l)}$ ;
        \IF{$\nabla g_v^{(l)} > \tau_{\text{add}}^{(l)}$ (Eq.(\ref{eq:12}))}
            \STATE Spawn Gaussians at scale $l$;
            \STATE Activate deformation at level $l+1$;
        \ELSE
            \STATE break;
        \ENDIF
    \ENDFOR
\ENDFOR
\FOR{each anchor $v \in V$}
    \STATE Apply redundant Gaussian removing using Eq.(\ref{eq:13}) for masked rendering;
\ENDFOR
\STATE Optimize total loss using Eq.(\ref{eq:14});
\RETURN Updated $\mathcal{G}_t$.
\end{algorithmic}
\end{algorithm}

The forward masking process selects dynamic anchors by distinguishing motion across adjacent multi-view frames. Specifically, temporal variations between two consecutive frames enable a coarse estimation of spatial motion for individual image pixels. The motion is then back-projected into 3D space to identify spatial position with significant displacement. If an anchor's coverage consistently exhibits prominent motion patterns over multiple frames, it is designated as a dynamic anchor,  which subsequently guides the optimization of dynamic objects.

The backward masking is designed to select a subset of camera views aligned with these dynamic regions. We introduce a relevance score for each camera view $c_k$. Let $\mathrm{IoU}(c_k, v)$ denote the normalized intersection-over-union between the image-space projection of anchor $v$ and the field-of-view of camera $c_k$. Given a threshold $\tau_{\text{view}}$, we define the view relevance score as
\begin{equation}
S(c_k) = \sum_{v\in V} \mathbf{1}\left[\mathrm{IoU}(c_k, v) > \tau_{\text{view}}\right] \cdot \omega(c_k, v), \label{eq:15}
\end{equation}
where $\mathbf{1}()$ is the indicator function and $\omega(c_k, v)$ is a direction weight, defined as
\begin{equation}
\omega(c_k, v) = |\mathbf{n}_v^\top \mathbf{d}_{c_k}|, \label{eq:16}
\end{equation}
where $\mathbf{n}_v$ is the average normal vector of Gaussians in anchor $v$, $d_v$ denotes the normal direction of the viewpoint $d_{c_k}$ . 

Given the ranking of \( S(c_k) \), the top-ranked views are selected for training, as they are considered the most relevant to the dynamics. This bidirectional masking mechanism reduces computational redundancy caused by uninformative viewpoints of dynamic scenes.
\vspace{-3mm}
\subsection{Algorithmic Summarization}
\label{3.6}


We summarize
~\M~dynamic scene optimization approach in Algorithm~\ref{alg:dynamic_update}, which performs multi-scale guided dynamic updates through hybrid deformation-spawning optimization (\ref{3.3}) combined with bidirectional adaptive masking (\ref{3.5}).

The algorithm operates on the pre-initialized multi-resolution Gaussian hierarchy, applying the hybrid deformation-spawning optimization strategy across multiple resolution levels while using bidirectional adaptive masking to focus computation on dynamic regions and informative views. This approach ensures efficient and targeted optimization for dynamic scene reconstruction while maintaining scale-aware structure throughout the process.

\noindent \textbf{Implementation Details.} 
All experiments are conducted in a virtual environment using Python 3.9 and PyTorch 2.1.0 as the primary deep learning framework. Additional libraries include plyfile 0.8.1 for data handling, torchaudio 0.12.1 for audio processing, and torchvision 0.13.1 for computer vision tasks.
Training and inference were performed on a workstation equipped with an NVIDIA RTX 4090 GPU with CUDA 12.6 support, complemented by cudatoolkit 11.8 for GPU acceleration. The proposed method is implemented in PyTorch and trained under the above configuration, ensuring seamless integration of hardware capabilities and software functionalities for reliable experimental outcomes.

\begin{figure*}[!t]
	\centering
	\includegraphics*[width=1\linewidth]{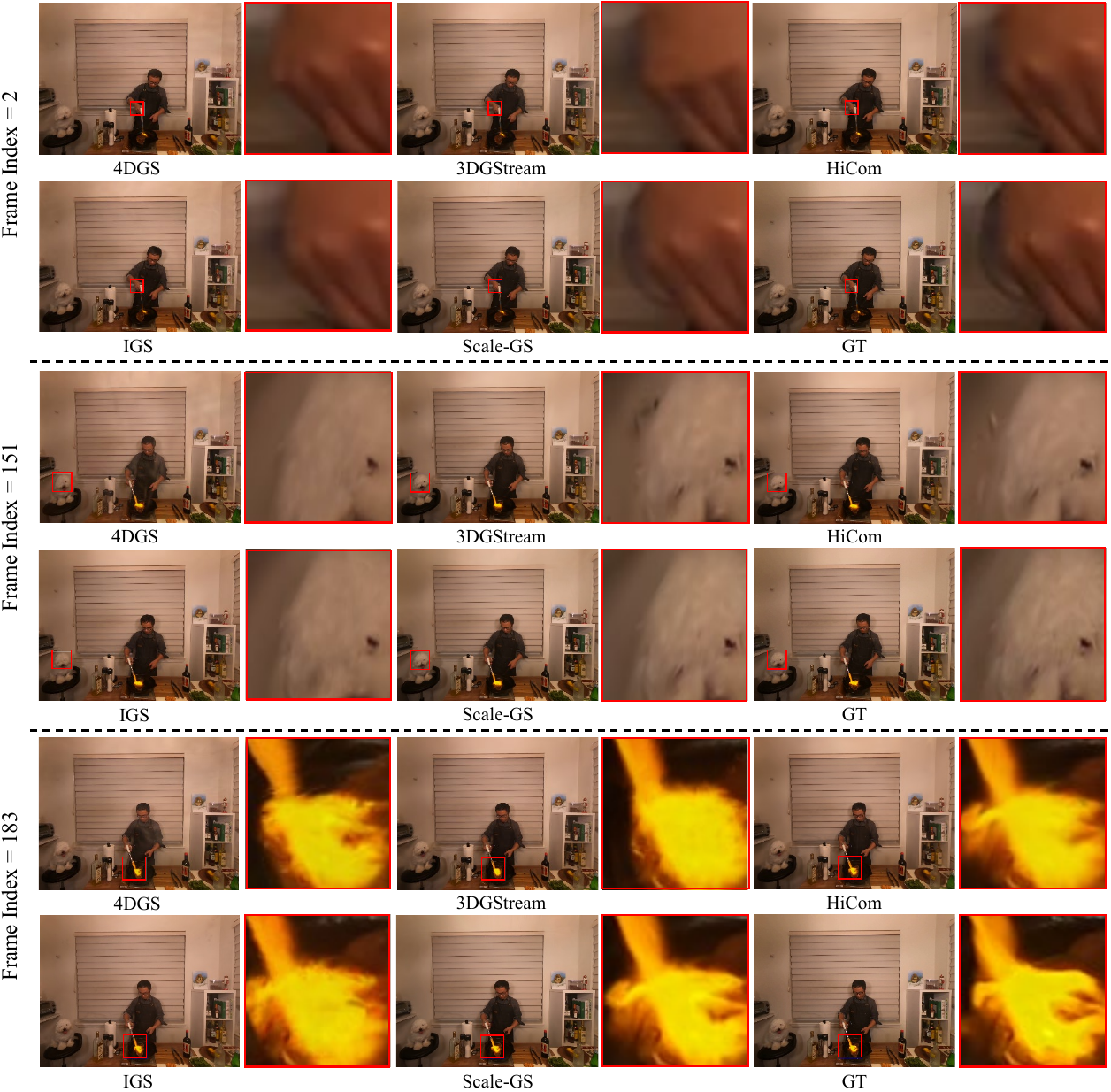}
	\caption{Qualitative comparison results on the NV3D datasets(scene flame-steak). The frame index is 2, 151, and 183 from up to down. For each frame index, we present the result of 4DGS, 3DStream, HiCom, IGS, \M~, and ground truth.}
    \label{nv3d}
\end{figure*}

\section{Experiment Results}
\label{sec4}
In this section, we thoroughly analyze the experiment results of~\M. Specifically, we first introduce the dataset and the experimental setup. Later, we present the qualitative results and quantitative results. Finally, we conduct the ablation study to illustrate the effectiveness of each key module. 

\subsection{Datasets}

We evaluate ~\M~method on three real-world dynamic scene datasets: the MeetRoom dataset\cite{li2022streaming}, the NV3D (Neural 3D Video) dataset\cite{li2022neural}, and the Google Immersive Light Field Video dataset\cite{broxton2020immersive}. All of the datasets exhibit complex motion and occlusion patterns, thus suitable for evaluating the free-viewpoint rendering.

\begin{figure*}[!t]
	\centering
	\includegraphics*[width=1\linewidth]{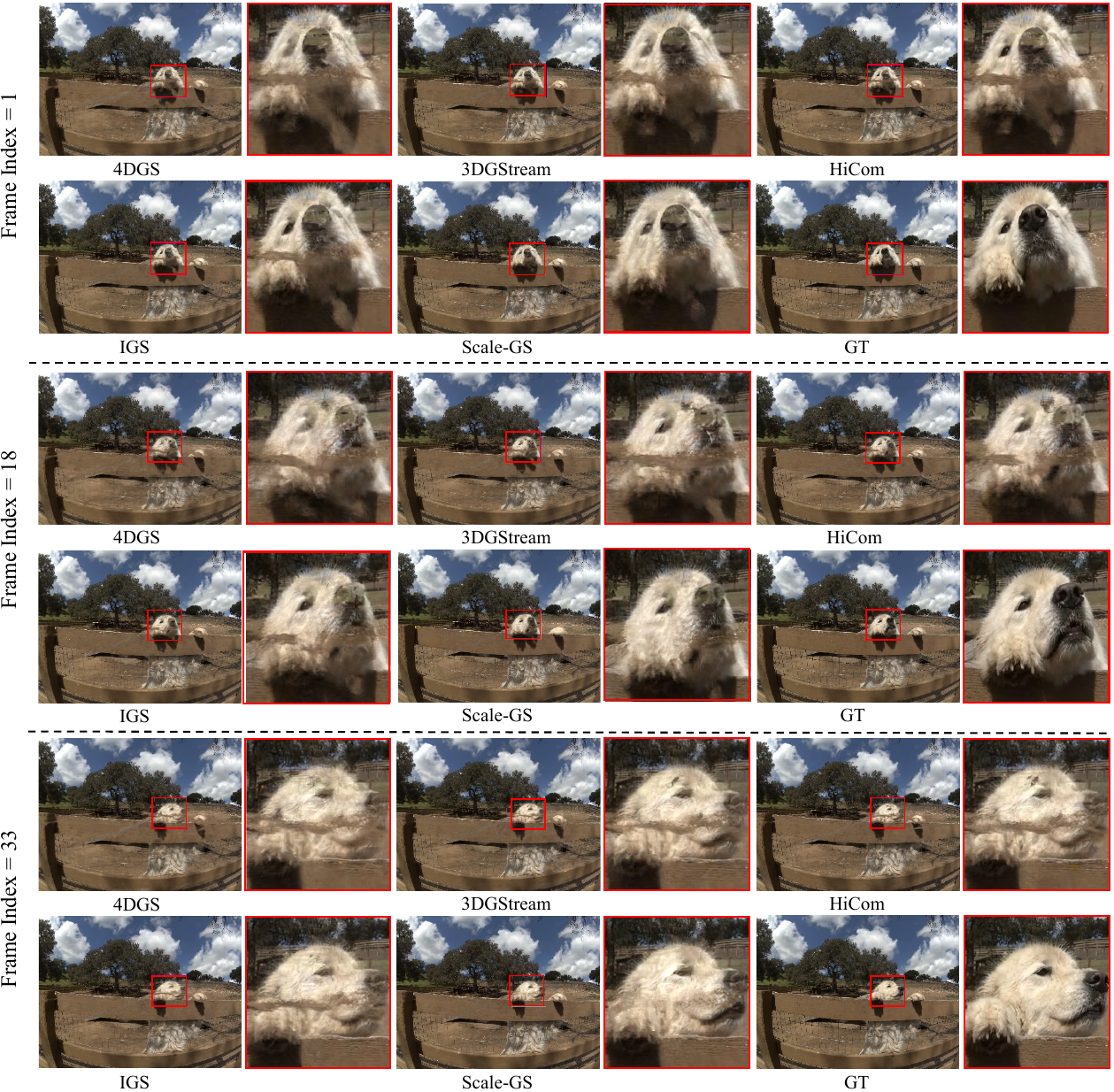}
	\caption{Qualitative comparison results on the Google Immersive datasets(scene Dog). The frame index is 1, 18 and 33 from up to down. For each frame index, we present the result of 4DGS, 3DStream, HiCom, IGS, \M~, and ground truth.}
    \label{immersive}
\end{figure*}

\noindent \textbf{NV3D Dataset\cite{li2022neural}.}~The Neural 3D Video dataset contains six dynamic scenes captured using a synchronized 21-camera array arranged in a semi-circular configuration. Each camera records the frames at a resolution of $2704 \times 2028$ with 300 frames, including various human motion interactions under indoor lighting conditions.

\noindent \textbf{MeetRoom Dataset\cite{li2022streaming}.} The MeetRoom dataset provides 3 indoor dynamic scenes recorded with 13 synchronized cameras, each capturing at $1280 \times 720$ resolution with 300 frames. The subjects are performing structured activities such as sitting, walking, or conversing in an office-like setting.

\noindent \textbf{Google Immersive Dataset\cite{broxton2020immersive}.} This dataset contains 15 complex dynamic scenes captured using a high-fidelity immersive light field video rig consisting of 46 time-synchronized cameras on a 92 cm diameter hemisphere, each capturing at $2560 \times 1920$ resolution. The system supports a large viewing baseline (up to 80 cm) and a wide field of view ($>$220°), posing very challenging novel view synthesis scenes.

\subsection{Experimental Setup}

\noindent \textbf{Hyper-parameter settings.}
In the qualitative experiment and quantitative experiment, we adopt a three-level hierarchical structure ($L=3$), corresponding to an image resolution pyramid obtained by downsampling the original input to $\{1/4, 1/2, 1\}$ for levels $l = 1, 2, 3$, respectively.
We set $\lambda_{\text{SSIM}} = 0.2$, $\lambda_{\text{r}}=0.001$ and the level-specific gradient thresholds $\tau_{\text{add}}^{(l)} = 0.01 / 4^{l-1}$, which yields $\{\tau_{\text{add}}^{(l)}\} = \{0.01, 0.0025, 0.000625\}$ for levels $l = 1, 2, 3$ respectively.

\noindent\textbf{Metrics}.
We assess the rendering fidelity of ~\M~method using three widely adopted perceptual and photometric metrics: PSNR (Peak Signal-to-Noise Ratio), SSIM (Structural Similarity Index), and LPIPS (Learned Perceptual Image Patch Similarity). These metrics respectively reflect pixel-level accuracy, structural coherence, and perceptual similarity. Furthermore, we compare the training time and rendering time to specify the efficiency of the proposed~\M. 

\noindent\textbf{Baselines.}
We evaluate ~\M~against five state-of-the-art dynamic scene rendering methods that represent different approaches to temporal modeling and real-time performance.

Deformable 3D Gaussians~\cite{yang2024deformable} extends the static 3D Gaussian splatting framework to handle dynamic scenes through a canonical space representation. The method learns a set of 3D Gaussians in canonical space and captures temporal variations using a deformation field implemented as an MLP that predicts position, rotation, and scaling offsets. To address potential jitter from inaccurate camera poses, the authors introduce an annealing smooth training mechanism, enabling both high-fidelity rendering quality and real-time performance.

4DGS~\cite{wu20244d} takes a different approach by incorporating a Gaussian deformation field network that operates on canonical 3D Gaussians. The method employs a spatial-temporal structure encoder coupled with a multi-head deformation decoder to predict Gaussian transformations across time. By modeling both Gaussian motion and shape changes through decomposed neural voxel encoding, 4DGS achieves efficient real-time rendering while maintaining temporal consistency.

3DGStream~\cite{sun20243dgstream} focuses on streaming applications, enabling on-the-fly training for photo-realistic free-viewpoint videos. The method introduces a Neural Transformation Cache (NTC) to model 3D Gaussian transformations and employs an adaptive Gaussian spawn strategy for handling newly appearing objects. The framework operates through a two-stage pipeline: first training the NTC for existing Gaussians, then spawning and optimizing additional frame-specific Gaussians to accommodate emerging scene content.

HiCoM~\cite{gao2024hicom} presents a comprehensive framework specifically designed for streamable dynamic scene reconstruction. It combines three key components: a perturbation smoothing strategy for robust initial 3D Gaussian representation, a hierarchical coherent motion mechanism that captures multi-granular motions through parameter sharing within regional hierarchies, and a continual refinement process that evolves scene content while maintaining representation compactness.

IGS~\cite{yan2025instant} offers a generalized streaming framework centered around an Anchor-driven Gaussian Motion Network (AGM-Net). This network projects multi-view 2D motion features into 3D space using strategically placed anchor points to drive Gaussian motion. The method further incorporates a key-frame-guided streaming strategy that refines key frames and effectively mitigates error accumulation during long sequences.

\begin{table*}[!t]
  \centering
  \renewcommand{\arraystretch}{1.25}
  \caption{The quantitative results on different datasets. NV, MR, and GI are separately denoted  NV3D dataset, Meeting room dataset, and Google Immersive dataset. The best and second-best results are \fcolorbox{white}{lightred}{\textbf{red}} and \fcolorbox{white}{lightpurple}{purple}, respectively.}
  \vspace{-2mm}
  \setlength{\tabcolsep}{1.1mm}{ 
    \begin{tabular}{l|ccc|ccc|ccc|ccc|ccc} 
    \toprule
    & \multicolumn{3}{c|}{SSIM$_\uparrow$} & \multicolumn{3}{c|}{PSNR$_\uparrow$} & \multicolumn{3}{c|}{LPIPS$_\downarrow$} & \multicolumn{3}{c|}{Training time(s)$_\downarrow$} & \multicolumn{3}{c}{FPS$_\uparrow$} \\
    \midrule
    Method & NV & MR & GI & NV & MR & GI & NV & MR & GI & NV & MR & GI & NV & MR & GI \\
    \midrule
    Deform & 0.956 & 0.857 & 0.843 & 32.10 & 27.81 & 26.46 & 0.127 & 0.206 & 0.216 & 38    & 29  & 81    & 40    & 45    & 35 \\
    4DGS  & 0.959 & 0.870  & 0.853 & 32.23 & 28.69 & 27.88 & 0.109 & 0.184 & 0.214 & 8.2   & 7.5   & 73.5  & 30    & 38    & 26 \\
    3DG-S & 0.958 & 0.906 & 0.868 & 32.93 & 29.09 & 28.60 & \fcolorbox{white}{lightred}{\textbf{0.101}} & 0.145 & 0.209 & 9.6   & 7.7   & 76.1  & 210   & 252   & 190 \\
    HiCoM & \fcolorbox{white}{lightred}{\textbf{0.967}} & \fcolorbox{white}{lightred}{\textbf{0.909}} & 0.852 & 33.28 & 29.15 & 28.68 & 0.111 & 0.152 & 0.208 & 7.1   & 3.9   & 60.8  & \fcolorbox{white}{lightpurple}{256}   & \fcolorbox{white}{lightred}{\textbf{284}}   & \fcolorbox{white}{lightred}{\textbf{212}} \\
    IGS   & 0.965 & \fcolorbox{white}{lightred}{\textbf{0.909}} & \fcolorbox{white}{lightpurple}{0.909} & \fcolorbox{white}{lightpurple}{33.62} & \fcolorbox{white}{lightpurple}{30.13} & \fcolorbox{white}{lightpurple}{29.72} & 0.109 & \fcolorbox{white}{lightpurple}{0.143} & \fcolorbox{white}{lightred}{\textbf{0.168}} & \fcolorbox{white}{lightpurple}{3.6}   & \fcolorbox{white}{lightpurple}{3.2}   & \fcolorbox{white}{lightpurple}{43.2}  & 204   & 251   & 186 \\
    \M & \fcolorbox{white}{lightpurple}{0.966} & \fcolorbox{white}{lightpurple}{0.908} & \fcolorbox{white}{lightred}{\textbf{0.912}} & \fcolorbox{white}{lightred}{\textbf{34.47}} & \fcolorbox{white}{lightred}{\textbf{31.58}} & \fcolorbox{white}{lightred}{\textbf{31.18}} & \fcolorbox{white}{lightpurple}{0.106} & \fcolorbox{white}{lightred}{\textbf{0.142}} & \fcolorbox{white}{lightpurple}{0.169} & \fcolorbox{white}{lightred}{\textbf{3.2}}   & \fcolorbox{white}{lightred}{\textbf{3.0}}  & \fcolorbox{white}{lightred}{\textbf{37.3}}  & \fcolorbox{white}{lightred}{\textbf{274}}   & \fcolorbox{white}{lightpurple}{276}   & \fcolorbox{white}{lightpurple}{199} \\
    \bottomrule
    \end{tabular}%
  }
  \label{quan}%
  \vspace{-3mm}
\end{table*}%

\subsection{Qualitative Results}

We choose 2 scenes from NV3D and the Google Immersive dataset to conduct a comprehensive qualitative evaluation, including indoor scene and outdoor scene. 

\subsubsection{Comparison in Indoor Scene}
Fig.~\ref{nv3d} presents the visual result on the NV3D dataset on 3 progressive frame index. In this scenario, it is critical to analyze the motion of the human and the dog and accurately reconstruct the dynamics of the flame. According to Fig.~\ref{nv3d}, \M~reconstructs significantly sharper and accurate fine-grained details compared to the baselines.

Specifically, in the frame 2, the 4DGS and 3DGStream fail to capture the trailing motion of hands, exhibiting noticeable blur. Meanwhile, four baselines render blurred occluded backgrounds, showing a large difference from the ground truth. The proposed~\M~not only reconstructs the clear texture of the blender, but renders the contour of the occluded object.
In the frame 151, 4DGS and IGS fail to properly render the dog's eyes. The 3DGStream and HiCom display obvious abnormal Gaussian points in the dog's head area. Compared to the baselines, the~\M~renders a richly furred dog head without introducing floaters.
The frame 183 indicates that the~\M~accurately captures the natural shape variations of the flame and intricate details. Although the baselines achieve flame rendering from a global perspective, they do not match the ground truth in the zoomed-in regions.

\subsubsection{Comparison in Outdoor Scene}
Fig.~\ref{immersive}  presents outdoor comparative results on the Google Immersive dataset. Compared to indoor scenes, outdoor scenes cover a larger area, involve more rendering details, and are more challenging. The red bounding boxes highlight regions showing the dog's head. In addition to reconstructing the changes in the dog's facial features and head, it is necessary to render the complex fur.

According to Fig.~\ref{immersive}, the proposed~\M~reconstruct clear and consistent face of the dog, while the baselines fail to accurately represent the dog's facial contours. Specifically, 4DGS, 3DGStream, and HiCom exhibit strong blurring around the dog's nose area in the frame 18. Moreover, the 3DGStream and HiCom methods display abnormal Gaussian representations at the dog's eyebrows.
In contrast, the subtle details around the eyes and nose of the~\M~are preserved with high fidelity, especially maintaining consistent quality across different viewing angles.

Overall, Fig.~\ref{nv3d} and Fig.~\ref{immersive} demonstrate that~\M~achieves superior visual quality and faithfulness compared to the baselines, particularly in challenging scenarios involving complex human interactions, dynamic elements like flame, and intricate textures such as animal fur.

\subsection{Quantitative Results}
Tab.~\ref{quan} presents the quantitative comparison to baselines on the 3 different datasets. We evaluate the novel view quality and rendering efficiency. To ensure a fair comparison, all competing methods are evaluated using the same set of Gaussians initialized from the $0$-th frame, and the same variant of Gaussian splatting rasterization is applied consistently across all approaches.



According to Table~\ref{quan},~\M~outperforms the baselines and achieves the best reconstruction quality with the fastest training speed. In the rendering quality aspect,~\M~achieves a PSNR improvement from 33.62dB (second-best IGS) to 34.47dB on NV3D dataset, from 30.13dB to 31.58dB on Meeting Room dataset, and from 29.72dB to 31.18dB on Google Immersive dataset, demonstrating consistent quality enhancement across all evaluation scenarios. For SSIM metrics,~\M~achieves competitive performance with 0.912 on Google Immersive dataset (best), 0.966 on NV3D dataset (0.001 below the best), and 0.908 on Meeting Room dataset (0.001 below the best), indicating excellent structural preservation. The LPIPS scores show perceptual quality improvements, with ~\M~achieving the best result of 0.142 on Meeting Room dataset and maintaining competitive performance on other datasets.

The sub-optimal performance of competing methods can be attributed to several limitations: (1) Global deformation inefficiency: Methods like Deform and 4DGS apply deformation to all Gaussians uniformly, leading to unnecessary computations on static regions and reduced optimization focus on truly dynamic areas. (2) Single-scale representation constraints: Traditional approaches like 3DG-S and HiCoM operate at fixed resolutions, failing to leverage hierarchical optimization strategies. In contrast, our multi-scale approach enables rapid convergence by first focusing on coarse-scale Gaussians that capture major scene changes, then progressively refining finer-scale Gaussians with increasing precision. This hierarchical refinement allows the optimization to quickly identify and target only the Gaussians that require updates at each scale level, resulting in accelerated training convergence and progressively improved rendering quality as finer details are incorporated. (3) Lack of adaptive supervision: Existing methods fail to adaptively select the most informative views and spatial regions during training, resulting in suboptimal resource allocation and slower convergence. In contrast, our multi-scale framework with hybrid deformation-spawning policy and bidirectional adaptive masking addresses these limitations systematically, enabling both superior reconstruction quality and computational efficiency.

\subsection{Ablation Study}

\subsubsection{The Evaluation of Scale Number}
We conduct a group of ablation study to investigate the influence of the scale number of~\M. We choose the Face Print scene from the Google Immersive Dataset and set the scale to 2, 3, 4, and 5, respectively. The result of average SSIM, PSNR, LPIPS, and training time is demonstrated in Table~\ref{multi-scale ablation}. 

The experimental results indicate that the configuration with 3 scales achieves the best performance across all metrics, yielding an SSIM of 0.916, a PSNR of 31.233 dB, and an LPIPS score of 0.154. 
It is noted that with the increase of the scale number, the rendering quality may decline under the same training iterations.
This is because the redundant scale requires training computation, thus the convergence of each scale becomes insufficient. Therefore, 3 level of scales strike the tradeoff between visual quality and training efficiency.

\begin{figure}[!t]
	\centering
	\includegraphics*[width=0.9\linewidth]{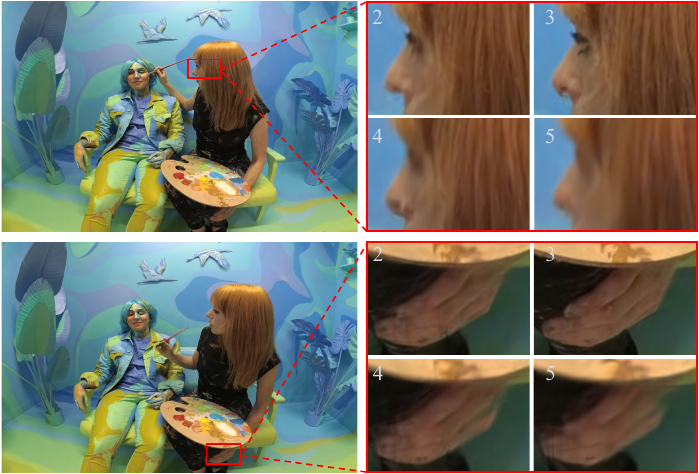}
	\caption{The visual quality comparison with different numbers of scales. From top-left to bottom-right: results with 2, 3, 4, and 5 scales, respectively.}
        \label{level}
        \vspace{-3mm}
\end{figure}

To further validate the visual performance gap across different scale settings, in Fig.~\ref{level}, we present a qualitative comparison of the zoom-in region in the same scene. It can be observed that the 3-scale configuration produces the most faithful reconstruction, particularly in facial regions such as the eyes, eyelashes, and hair strands, closely resembling the ground truth. The hand region near the drawing board is reconstructed the clearest details with the scale number as 3. In comparison, other scale configurations exhibit varying degrees of Gaussian transparency or blurring in the hand area, which negatively affects the rendering quality.

\begin{table}[!t]
  \centering
  \caption{Ablation study of scale number on Google Immersive Dataset, where time indicates training time.}
  \vspace{-2mm}
  \setlength{\tabcolsep}{13pt} 
  \begin{threeparttable}
  \begin{tabular}{c c c c c}
    \toprule
    Scale & SSIM$_\uparrow$ & PSNR$_\uparrow$ & LPIPS$_\downarrow$ & Time(s)$_\downarrow$ \\
    \midrule
    2     & 0.893 & 30.602 & 0.168 & 36 \\
    \rowcolor{lightgray!40} 3     & \textbf{0.916} & \textbf{31.233} & \textbf{0.154} & \textbf{33} \\
    4     & 0.881 & 29.917 & 0.170  & 35 \\
    5     & 0.857 & 28.242 & 0.191 & 39 \\
    \bottomrule
  \end{tabular}
\begin{tablenotes}
        \footnotesize
       \item[*]
       Our setting is highlighted with a~\colorbox{lightgray!40}{shaded background}.
       \vspace{-3mm}
      \end{tablenotes}
      \end{threeparttable}
  \label{multi-scale ablation}
\end{table}
\subsubsection{The Effectiveness of Hybrid Deformation-spawning}
To evaluate the effectiveness of the hybrid deformation-spawning strategy, we conduct an ablation study on the MeetRoom dataset by comparing three variants: (1) the full hybrid strategy, (2) w/o deformation, and (3) w/o spawn. The results, summarized in Table~\ref{deform-spawn ablation}, show that the hybrid approach achieves the best performance across all evaluated metrics, in which SSIM is 0.913, PSNR is 31.602 dB, and LPIPS is 0.138, and the least training time per frame (2.9 seconds). In contrast, removing either component leads to a noticeable degradation in both reconstruction quality and efficiency.

\begin{figure}[!t]
	\centering
	\includegraphics*[width=1\linewidth]{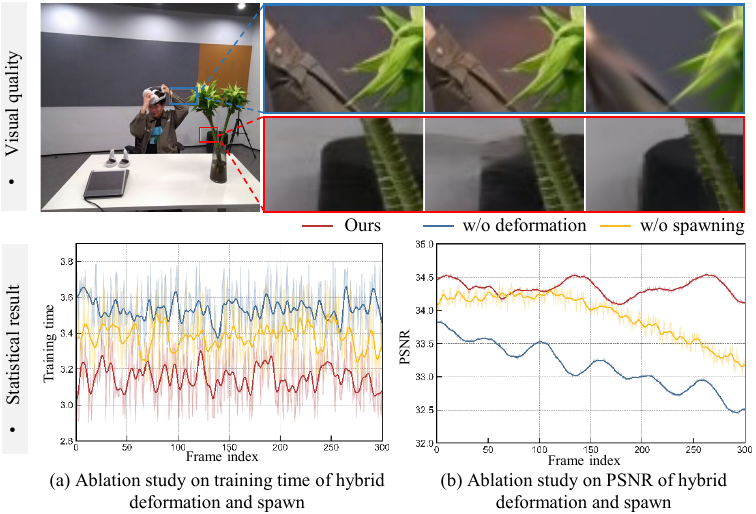}
	\caption{The visual quality(from left to right: full hybrid, w/o deformation, and w/o spawning) and statistical results of the ablation study on hybrid deformation-spawning in the Meeting Room dataset. }
        \label{line}
\end{figure}

Furthermore, we demonstrate the visual quality(top) and the statistical PSNR and training time(bottom) comparison among the full hybrid strategy, without deformation, and without spawning in Fig.~\ref{line}. According to the visual result, the full hybrid strategy exhibits clear edge and accurate color representation. Moreover, the details of the plant leaves and the fabric textures are also well preserved. These visual results further validate that the hybrid mechanism achieves high-quality rendering and efficient training of dynamic scenes.

As shown in Fig.~\ref{line}, our hybrid approach consistently achieves the shortest training time per frame (averaging around 3.2 seconds), while both ablated variants require significantly longer training periods. Specifically, the “w/o spawn” variant, though taking more time than our method, has relatively more stable training duration compared to “w/o deformation”; the “w/o deformation” variant shows considerable training time fluctuations throughout the sequence and exhibits the highest training costs. Fig.~\ref{line}(b) demonstrates the PSNR evolution across frames, where our method maintains relatively stable and high PSNR values (around 34.5 dB). The “w/o spawn” variant can keep a decent level of PSNR for a while but shows more pronounced fluctuations than our method as frames progress, while the “w/o deformation” variant experiences a more gradual yet evident decline in PSNR over time and overall lower reconstruction quality. The results demonstrate that the \M~method not only trains faster but also maintains more stable rendering quality throughout the sequence. 

Even the dynamic nature of the scene leads to unstable PSNR variations, the hybrid deformation and spawning strategy performs the best in both visual quality and training efficiency in most cases. This suggests that the joint optimization of deformation and spawning enables more efficient and stable learning of dynamic scene representations.


\subsubsection{Viewpoint Selection}

To validate the effectiveness of Bidirectional Adaptive Masking (BAM) mechanism, we conduct ablation studies on five challenging scenes from the Google Immersive Dataset: Car, Goats, Dogs,  Face Paint, and Welder. These scenes feature diverse dynamic content ranging from fast-moving objects to complex deformations, providing a comprehensive testbed for evaluating view selection strategy.

\begin{table}[!t]
  \centering
  \caption{Ablation study of the hybrid deformation-spawning policy, where time indicates training time.}
  \vspace{-3mm}
  \setlength{\tabcolsep}{10pt} 
  \begin{tabular}{c c c c c}
    \toprule
     & SSIM$_\uparrow$ & PSNR$_\uparrow$ & LPIPS$_\downarrow$ & Time(s)$_\downarrow$ \\
    \midrule
    \rowcolor{lightgray!40}deform + spawn & \textbf{0.913} & \textbf{31.602} & 0.138 & \textbf{2.9} \\
    w/o deformation & 0.854 & 26.719 & 0.245 & 4.6 \\
    w/o spawning & 0.887 & 29.304 & \textbf{0.127} & 3.8 \\
    \bottomrule
  \end{tabular}
  \label{deform-spawn ablation}
  \vspace{-4mm}
\end{table}

The quantitative results are presented in Table \ref{view-select}. ~\M~BAM-based view selection consistently outperforms the baseline across all tested scenes, achieving improvements of 0.7-1.7 dB in PSNR and 0.007-0.035 in SSIM. The most significant improvement is observed in the Welder scene, where BAM selection achieves 30.281 dB PSNR compared to 28.552 dB without selection, representing a substantial 1.729 dB gain. This scene benefits particularly due to its complex welding sparks and rapid lighting changes, where targeted view selection effectively focuses learning on the most informative perspectives.

The improvement across diverse scene types demonstrates that BAM successfully identifies and prioritizes views that provide meaningful supervision for dynamic regions. By filtering out redundant or less informative viewpoints, the BAM not only improves reconstruction quality but also enhances training efficiency. The bidirectional masking—selecting both spatial anchors and camera views—proves essential for handling the complexity of multi-view dynamic scene reconstruction.

These results validate that bidirectional view selection mechanism effectively ensure that Gaussians receive supervision from the most relevant camera perspectives, leading to more accurate and stable dynamic scene modeling.

\vspace{-3mm}
\section{Conclusion and Future Work.}
\label{sec5}
\noindent\textbf{Conclusion.} In this work, we present \M, a scalable Gaussian Splatting framework for efficient and redundancy-aware dynamic scene training. 
The proposed~\M~proposes the anchor-based  multi-scale Gaussian representation
integrating a hybrid deformation–spawning optimization, redundant Gaussian removing, and a bidirectional adaptive masking module.
Our method effectively reduces the computational overhead of existing approaches by filtering static or irrelevant Gaussian spheres. The hybrid deformation–spawning strategy preserves the structured inference from deformation while enhancing the model’s capacity to represent large-scale dynamic motions. Extensive experiments demonstrate that~\M~achieves superior visual fidelity and significantly accelerates training.  The proposed framework opens up promising opportunities for remote immersive experiences such as VR, AR, and immersive video conferencing.

\noindent\textbf{Future work.} To further reduce the redundant computation and storage, we aim to incorporate semantic understanding into the multi-scale representation to improve the representation efficiency of Gaussian spheres.
The semantic information guides the allocation and prioritization of Gaussian spheres by adapting their scale, density, and training frequency according to the semantic importance and structural complexity of different areas. 
Furthermore, integrating~\M~with neural compression could further enhance scalability and enable deployment in bandwidth and resource-constrained environments.

\begin{table}[!t]
\vspace{-2mm}
    \centering
    \caption{The ablation study of viewpoint selection results, where time indicates training time.}
    
    \renewcommand{\arraystretch}{1.1}
    \label{view-select}
    \setlength{\tabcolsep}{3.5mm}{	
    \begin{tabular}{lcc|cc}
        \toprule
        & \multicolumn{2}{c|}{\fcolorbox{white}{lightgray!40}{BAM selection}}& \multicolumn{2}{c}{w/o BAM selection} \\
       \cmidrule(lr){2-3} \cmidrule(lr){4-5}
        Scene& PSNR$_\uparrow$ & SSIM$_\uparrow$ & PSNR$_\uparrow$ & SSIM$_\uparrow$ \\
        \midrule
        Car & 31.320$_\text{+1.012}$ & 0.917$_\text{+0.011}$ & 30.308 & 0.906 \\
        Goats & 31.778$_\text{+1.150}$ & 0.922$_\text{+0.007}$ & 30.628 & 0.915 \\
        Dogs & 31.022$_\text{+0.737}$ & 0.914$_\text{+0.021}$ & 30.285 & 0.893 \\
        Paint & 31.751$_\text{+0.906}$ & 0.916$_\text{+0.006}$ & 30.845 & 0.910 \\
        Welder & 30.281$_\text{+1.729}$ & 0.901$_\text{+0.035}$ & 28.552 & 0.866 \\
        \bottomrule
    \end{tabular}}
    
\end{table}

\vspace{-3mm}
\bibliographystyle{IEEEtran}
\bibliography{ref}

\end{document}